\documentclass[journal]{IEEEtran}
\usepackage{ifpdf}
\usepackage{cite}
\usepackage{multirow}
\usepackage{array}
\ifCLASSINFOpdf
\usepackage[pdftex]{graphicx}
\usepackage{epstopdf}
\DeclareGraphicsExtensions{.pdf,.png,.jpg,.tif}
\usepackage[cmex10]{amsmath}
\usepackage{algorithmic}
\usepackage{array}
\usepackage{enumitem}
\ifCLASSOPTIONcompsoc
  \usepackage[caption=false,font=normalsize,labelfont=sf,textfont=sf]{subfig}
\else
  \usepackage[caption=false,font=footnotesize]{subfig}
\fi
\usepackage{fixltx2e}
\usepackage{amsfonts}
\usepackage{dblfloatfix}
\ifCLASSOPTIONcaptionsoff
  \usepackage[nomarkers]{endfloat}
 \let\MYoriglatexcaption\caption
 \renewcommand{\caption}[2][\relax]{\MYoriglatexcaption[#2]{#2}}
\fi
\usepackage{url}
\usepackage{xspace}
\usepackage[linesnumbered,ruled,vlined]{algorithm2e}
\usepackage{algorithmic}
\usepackage[switch]{lineno} 
\usepackage{lipsum}

\ifCLASSOPTIONpeerreview
 \begin{center} \bfseries EDICS Category: 3-BBND \end{center}
\fi
\IEEEpeerreviewmaketitle

\begin{document}
\title{Occlusion-Aware Human Pose Estimation with Mixtures of Sub-Trees}
\author{Ibrahim Radwan$^{*}$, Abhinav Dhall and~Roland Goecke
\IEEEcompsocitemizethanks{\IEEEcompsocthanksitem * Corresponding author \protect\\Ibrahim Radwan, Abhinav Dhall and Roland Goecke are with the University of Canberra, Bruce ACT, Australia.\protect\\
E-mail: \{ibrahim.radwan,~abhi.dhall\}@canberra.edu.au, roland.goecke@ieee.org}
\thanks{}}
\markboth{}%
{Radwan \MakeLowercase{\textit{et al.}}: Occlusion-Aware Human Pose Estimation with Mixtures of Sub-Trees}
\maketitle

\begin{abstract}
In this paper, we study the problem of learning a model for human pose estimation as mixtures of compositional sub-trees in two layers of prediction. This involves estimating the pose of a sub-tree followed by identifying the relationships between sub-trees that are used to handle occlusions between different parts. The mixtures of the sub-trees are learnt utilising both geometric and appearance distances. The Chow-Liu (CL) algorithm is recursively applied to determine the inter-relations between the nodes and to build the structure of the sub-trees. These structures are used to learn the latent parameters of the sub-trees and the inference is done using a standard belief propagation technique. The proposed method handles occlusions during the inference process by identifying overlapping regions between different sub-trees and introducing a penalty term for overlapping parts. Experiments are performed on three different datasets: the Leeds Sports, Image Parse and UIUC People datasets. The results show the robustness of the proposed method to occlusions over the state-of-the-art approaches.
\end{abstract}


\section{Introduction}
\label{MOST::sec_introduction}
\IEEEPARstart{T}{his} paper addresses the problem of articulated human pose estimation in real-world images. Human pose estimation is a long-standing and challenging problem in computer vision. It is an important building block for applications such as human activity recognition, attribute recognition and human-computer interaction. However, estimating the 2D pose from real-world images is a challenging task due to the high variability in human body poses, the large degree of freedom and the hallucination of body limbs. Many algorithms have been proposed to cope with these different challenges and also to improve the accuracy of the estimated poses. The key idea in most of these methods is to describe the appearance of human body as well as the relationship between the body parts in a specific pose.

A mixture of sub-trees (MST) style algorithm for pose estimation is developed in this paper. The algorithm builds on and improves a popular ``flexible mixture-of-parts'' (FMP) approach introduced by Yang and Ramanan \cite{YiYang11}. The FMP approach represents the body pose using a tree-structured part-based model. Model parts in FMP typically correspond to body joints or mid-points of body parts. Appearance of each part is modelled with a collection of HOG templates. Parts are combined using soft constraints on their mutual position and compatibility constraints on types of adjacent part templates. Significant limitations of the FMP model are that (1) it does not encode higher order relationships between parts, (2) tree-structured pairwise connections between parts are defined in an ad-hoc fashion to follow kinematic relationships between parts, and (3) the approach fails to handle body part occlusions.

In Wang and Li \cite{FWang13}, encoding the non-physical connections has been addressed by learning combined parts on top of the single parts. However, learning combined parts only exploits the appearance context of the parts. That is, the combined parts can improve the scores of an occluding part such that it is in the same location as an occluded part. This would confuse the inference in cases of occlusion just as in the FMP method. An improvement may be achieved via utilising more appearance features, which can discriminate between the different body parts, at the expense of increased complexity. In any case, relying on the appearance features in the occluded region of the body parts will not be useful. The solution proposed here is to employ both the discriminant appearance features as well as the configuration parameters between the body parts and their neighbours. When the model is a tree structure, finding an analytic and exact solution is applicable as the local parts are independent. However, in the presence of occlusion, these parts are no longer independent and new edges between the occluded and occluding parts should be (temporarily) established to represent this relationship. This would result in a non-tree structured model, thus performing the standard message passing and achieving exact inference is intractable.

The mixtures of sub-trees based model addresses these limitations. MST introduces mid-level components that include several adjacent body parts and are constructed to encode specific part appearance and geometric configuration information. Connectivity of parts within each component is determined from training data. The proposed model is structured such that it builds the parent-child relationships according to the correlation between the observed features of the body parts. That is, the most correlated parts would result in establishing an edge between them. Building the structure of the model based on the appearance or geometric similarities gives more flexibility to combine the occluded and the occluding regions in one sibling group. Following this scenario, we clustered the nodes, which belong to each limb, into different mixtures. Each mixture will have a unique sub-tree structure, which results in mixtures of sub-trees (MST) for different limbs.

The motivation behind learning multiple mixtures with different structures for each limb is to provide flexibility for the connections between the parts that belongs to each limb. The MST model also incorporates an additional term that takes self-occlusion into account. This term down-weights a hypothesis of a part that has substantial overlap with a strong hypothesis for other parts. Such down-weighting encourages solutions without over-lapping parts. Technically, the MST model produces a way to localise and estimate the candidate positions of the limb in early stages via implementing non-maximum suppression on the score maps of the limb's mixtures. This helps predicting the parts with occlusion by testing the overlapping region between the candidates of the limbs.  The occlusion handling step in the MST model improves on the rectification methods developed in \cite{RadwanICME2012}. MST incorporates \emph{occlusion reasoning} directly into the inference procedure, whereas the rectification methods in \cite{RadwanICME2012} aim to fix incorrect pose estimation results as a post-processing step.

The proposed approach can be seen as stacking simple FMP models for subsets of body parts into a more complex FMP model. Alternatively, it can also be seen as introducing a ``poselet'' type representation \cite{BourdevICCV2009} into the FMP model. The key contributions of the proposed method are as follows:
\begin{itemize}
\item Constructing a novel and more flexible tree structured model for estimating the 2D pose from monocular images, such that exact inference can be achieved.
\item Encoding the non-physical configurations via learning multiple mixtures with different sub-tree structures.
\item A self-occlusion reasoning method, which handles the presence of occlusion in testing images, even if the training images do not contain occlusions, thereby, eliminating the need to insert new edges in the tree model to represent these relationships.
\end{itemize}

%
%
\section{Related Work}
\label{MOST::sec_literaturereview}
The literature on human pose estimation is vast and varied in settings: applications range from highly-constrained MOCAP environments (e.g.\ \cite{Lan2005}) to extremely articulated baseball players (e.g.\ \cite{Mori2004}), to the recently popular `in the wild' datasets Buffy (from the TV show) and the PASCAL Stickmen (from amateur photographs) \cite{Ferrari2008}. Pose estimation methods can be categorised into three general types: 
\begin{enumerate}
\item Inferring poses from skeletal points. The major drawback of these methods is their need for a robust foreground segmentation process to extract the silhouette of the human body from the background.
\item Detecting poses by matching the input image template with a database of exemplars and finding the closest in the nearest neighbour sense. The limitation of these methods is their need for huge amounts of exemplars in the training phase.
\item Inferring the human pose firstly by detecting the human body parts and then learning the configuration parameters between these parts based on a prior model to estimate the relative order and connection of these parts. A successful framework to learn the configuration and connectivity parameters between the detected body parts is used in pictorial structure based pose estimation methods. 
\end{enumerate}
In this section, we focus on the recent and widely used Pictorial Structure (PS) models and methods for handling the occlusion problem.

%
%
\paragraph*{\textbf{Pictorial Structure-based Methods}}
While the Pictorial Structure idea dates back to Fischler \emph{et al.}\ \cite{Fischler:1973}, its great impact in recent years stems from Felzenszwalb \emph{et al.}\ \cite{Felzenszwalb2005}, who proposed to learn the deformation parameters between the body parts. Their approach is based on handling the relationship between the appearance of a body part and the compatibility with its adjacent parts being modelled as a cost function. The optimal solution is found by minimising the cost function. The main drawback of the approach lies in handling cluttered and dynamic background scenes containing highly deformable poses.

Ramanan \emph{et al.}\ \cite{Ramanan2005} employed the PS idea to find stylised poses, such as a walking person, by learning the difference between the background and foreground from consecutive frames and then tracking the person in consecutive video frames. This approach works well in videos for stylised poses with little occlusion in the parts.

Ramanan \emph{et al.}\ \cite{Ramanan2006} also proposed an iterative parsing process, which consists of two subsequent steps. Firstly, a deformable pose model is built based on an edge detector, which provides initial instances to the position of the articulated parts. Based on these initial instances, a region-based model is built for each part (and its background). The result of the region based model is the output for the body pose estimation. The drawback of this approach is that the whole process is relying upon a weak prior -- the response of edge detectors -- that may be very unrealistic in cluttered and dynamic backgrounds. 

Ferrari \emph{et al.}\ \cite{Ferrari2008} extended the work in \cite{Ramanan2006} by proposing a three-step framework: 1) Upper body detection. 2) Foreground highlighting using Grabcut \cite{Rother2004} to automatically learn foreground/background colour models from regions where the person is likely to be present/absent. This segmentation step removes the background clutter. 3) An edge based parsing step to find instances of the body parts. Based on these steps, the search space for the body parts is reduced and the problem of a weak prior is partially solved. However, the whole approach is relying strongly on the performance and accuracy of the upper body detector, which may be erroneous. Moreover, the self-occlusion problem remains still unsolved in this approach.

Ferrari \emph{et al.}\ \cite{Ferrari2009} extended their work in \cite{Ferrari2008} by exploiting two prior assumptions: 1) For the pose of the upper body, the orientation of the torso and head are near-vertical. This further reduces the search space for torso and head, thus improving the chances that they will be correctly estimated. 2) A so called repulsive model is used to partially solve the problem of occlusion only between the two arms. The model gives a high probability value when the two arms are not connected and a lower probability when they are connected. Although, this is a good solution when the problem of self-occlusion is limited to overlapping arms, that is not the case with articulated body parts with a large degree of freedom, where there are variants of self-occlusion types between the different body parts.

Felzenszwalb \emph{et al.}\ \cite{Felzenszwalb2008} proposed articulated pose estimation and action recognition using an object detection framework based on mixture of multi-scale deformable part models. This idea has become a building block in many recent, subsequent papers on human detection and pose estimation. The approach builds on the PS framework \cite{Felzenszwalb2005} by representing an object as a collection of parts arranged in a deformable configuration. This deformable configuration is represented by a spring like model, where this model is trained from images labelled with a bounding box around the object without knowing the exact position of the body parts. The part types are handled as latent variables and trained through a latent support vector machine (LSVM).

Andriluka \emph{et al.}\ \cite{Andriluka2009} also proposed a discriminative appearance model to overcome the problem of dynamic background and the need for the foreground segmentation step. Starting from the original PS method \cite{Felzenszwalb2005} for discriminatively detecting each body part, the normalised margin of each part is interpreted as the appearance likelihood for that part. Although this produces a general framework for both object detection and articulated pose estimation, the model is not able to estimate the human pose in highly occluded scenes.

One of the major challenges in pose estimation is the existence of large degrees of freedom and shortenings in the appearance of the parts. Most of the above mentioned literature utilises only one template as a detector for each part. However, Yang \emph{et al.}\ \cite{YiYang11} proposed a novel framework based on PS \cite{Felzenszwalb2005} using mixture of pictorial structures (MoPS) to encode the parts in different orientations and shortenings. This leads to mixtures of models to be used for estimating the pose of a given image. A shortcoming of the MoPS approach is the lack of handling non-physically connected parts, such as those that can occur between the two legs if they are overlapping, which leads in many cases to poor body part detections.

Tian \emph{et al.}\ \cite{TianZN12} proposed a flexible handling of the connection between the body parts through representing the parents as a set of latent nodes, which encode the types of the parts. The model is hierarchically structured and learnt in two layers. Due to the relationships between the observed and the latent nodes in the model being pre-defined, most of the non-physical connections have not been identified, which leads to poor part detections in cases of self-occlusion. In \cite{FWang13} and \cite{WangL13_ijcai}, the authors proposed a scheme to handle non-physical connections between parts. A recursive grouping algorithm is applied and a set of combined parts is learnt. The combined parts are utilised with the single templates of the body parts in a tree model to detect the human pose. 

In our proposed model, we follow the method of grouping the parts recursively into \emph{mixtures of sub-trees}, where each mixture has a structure of observed and latent nodes (similar to the combined parts in \cite{FWang13}) and each sub-tree is rooted in a latent node, which connects all mixtures together. The purpose of structuring the human body into mixtures of sub-trees is to encode non-physical connections between body parts and to utilise the roots of the sub-trees as stationary nodes for handling occlusion between sub-trees.

%
%
\paragraph*{\textbf{Occlusion-Reasoning Methods}}
Sigal and Black \cite{Sigal2006_short} modelled self-occlusion handling in the PS framework by imposing a set of constraints on the occluded and occluding regions, which are extracted after performing background subtraction, which renders it unsuitable for dynamic background scenes. Wang and Mori \cite{WangM08} constructed multiple tree models to estimate the pose and to reason the occlusion. In their method, the same strategy as in \cite{Sigal2006_short} is followed via assuming a known order of the body parts and re-weighting the scores of the occluded region through learning binary variables for the occluded and occluding parts. In our view, detecting the occlusion region requires estimating the candidates for the body parts and then re-weighting the scores of the occluded region different from the non-occluded parts. Moreover, reasoning the occlusion between all body parts is computationally expensive when the parts order is unknown and the parts order is impractical. In contrast, we propose an early detection phase for each part via propagating the messages through the parts in a sub-tree and handling the root of these sub-tree as stationary nodes, where the occlusion between the sub-trees is detected. Then, re-scoring of the regions with occlusion occurs.

The rest of the paper is organised as follows: The proposed method of modelling the structure of human poses as a mixture of sub-trees, the training and inference algorithms are presented in Sec.\ \ref{MOST::sec_MOST}. An extension to the proposed model for occlusion reasoning is proposed in Sec.\ \ref{MOST::sec_occlusion}. Sec.\  \ref{MOST::sec_experiments} provides quantitative and qualitative experimental results and a discussion of the proposed method compared with state-of-the-art approaches. Sec. \ref{MOST::sec_conclusion} concludes and summarises the findings of the paper.

%
%
\section{Modelling the Human Pose}
\label{MOST::sec_MOST}
In this section, the general steps of the proposed algorithm for modelling the body pose as mixtures of sub-trees are discussed. Beginning with how the model is initialised and structured based on the labels of the training data. Then, the training of this model is performed, followed by the prediction of the 2D pose for any input image through an inference step.
\subsection{Structuring a Sub-Tree}
Given an image $I$, the human pose can be depicted as a graph $G = (V,E)$, where the nodes $V$ represent the body parts and the edges $E$ are the links between the physically or ``non-physically" connected parts. To determine the parent-child relationships between the nodes, in the proposed model, let us start with the positions of 26 body parts as in \cite{YiYang11}. Each node \textbf{$v_i$} contains the position of the part $i$. The nodes $V$, which represent the body, are divided into sub-trees (limbs), with a sub-tree for each group of nodes for the following limbs: left arm, right arm, left leg and right leg. A sub-tree comprises set of nodes and different configurations. In addition, each sub-tree is rooted with a latent node, which will be used later to handle the occlusion relationships between the non-physically connected sub-trees. The position of the root node of each sub-tree can be at any point on the sub-tree space. In the experiments, the centre point of a sub-tree is selected to be the location of the root node.
\begin{figure*}[h]
\centering
\subfloat[\label{MOST::fig_structure_a}]{%
      \includegraphics[width=0.45\textwidth]{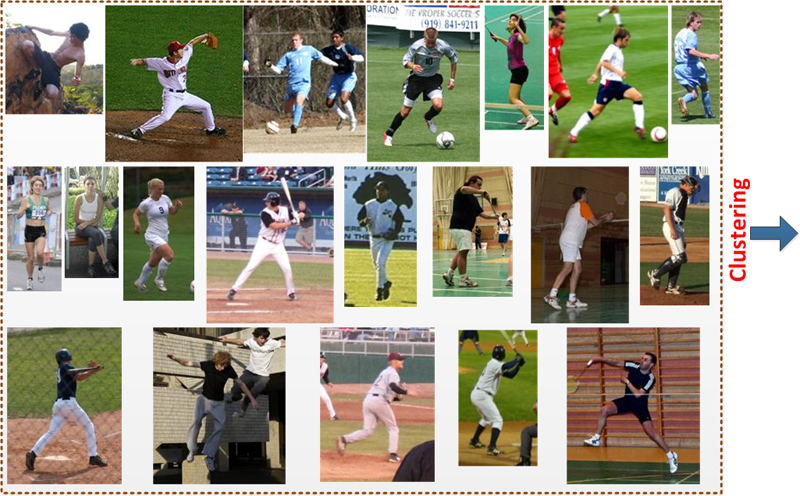}
    }
    \subfloat[\label{MOST::fig_structure_b}]{%
      \includegraphics[width=0.46\textwidth]{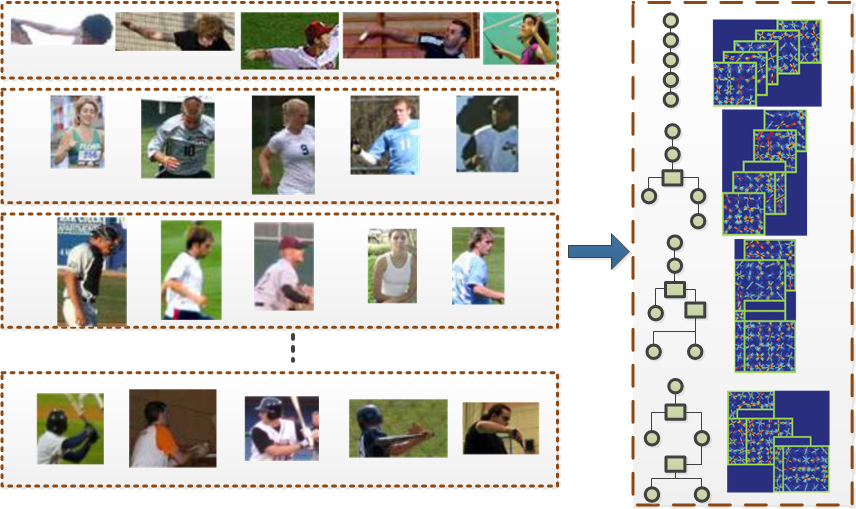}
    }
    \caption{Example of building a sub-tree's mixtures for the left arm: Starting with the input samples (a), these samples are categorised into different mixtures for the left arm where the correlating samples share similar appearance (first column in (b)). In the second column of (b), a sub-tree structure is constructed where the circles represent the observed nodes and the rectangles are the latent nodes. The structure of each mixture is here visualised with the corresponding HOG templates.}
\label{MOST::fig_subtree-structure}
\end{figure*}
%

To start with, the definition of sub-tree is provided as a collection of nodes/parts connected together on a limb. Each sub-tree in our model can be modelled by different mixtures, where each sub-tree's mixture shares similar geometric and appearance information. The mixtures of a sub-tree are rooted on a unified node. This root node is not a physical part of the body. However, it is a latent node, which is built from all of the messages, which go through the nodes of the sub-tree. This way of rooting the sub-trees on latent nodes helps investigating the self-occlusion, which is happening between the different limbs.

Now, to extract the structures of each sub-tree automatically, the appearance and the geometric distances between the nodes in a sub-tree and its root node have been concatenated to combine a vector for this sub-tree. This is done for all training data. The widely used K-means clustering is exploited to extract different mixtures $M$. The structure of each mixture is determined automatically based on the adjacency matrix, which is formed by computing the distance between features of node \textbf{$v_i$} and features of the root $v_0$ based on the Product-Moment Correlation Coefficient between those two nodes:
\begin{equation}
d(X,Y) = -\log( \dfrac{Cov({v_i},{v_0})} {\sqrt{Var({v_i})Var({v_0})}})
\end{equation}
where, $Cov(v_i, v_0) = n(\sum_{j=1}^{n} v_{i}^{j} v_{0}^{j}) - (\sum_{j=1}^{n} v_{i}^{j})(\sum_{j=1}^{n}v_{0}^{j})$ is the covariance between the feature vectors of node $v_i$ and its root $v_0$. $Var(v) = n(\sum_{j=1}^{n}(v^{j})^{2})-(\sum_{j=1}^{n}v^{j})^{2}$ is the variance of the feature vector of each node $v$.
 
Depending on the obtained distance matrix, the Chow-Liu grouping algorithm (\cite{Chow68}) is utilised recursively to construct different structures for the mixtures of the sub-tree. As in \cite{FWang13}, the CL-Grouping algorithm, proposed in \cite{Choi:2011}, is employed to determine the sibling groups between the body parts in the mixture of the sub-tree. This results in combined parts, which are then trained  to observe combined parts for each sibling group to handle the non-physical connections inside the same mixture. In our method, the CL-grouping is employed on the level of the mixtures of the sub-tree only, not the whole body. This results in identifying unique structures (mixtures) for each sub-tree. Moreover, the root of each sibling group is manipulated as a latent node. All mixtures of a sub-tree are rooted on a latent node, where this node is a transitional state in the model of the whole body. That is, the estimated scores of these nodes is utilised to measure the overlapping ratios between the sub-trees as in Section \ref{MOST::sec_occlusion}.
 
Given the geometric distances between the observed nodes in a sub-tree, the CL-grouping algorithm finds set of edges to connect these nodes from different potential structures by solving the optimisation problem:
\begin{equation}
T_{CL} = \operatorname*{\arg\,max}_{{T}\in \mathcal{T}} \sum_{(i,j) \in T} {\textit{I}(v_i;v_j)}
\end{equation}
where ${I(v_i;v_j)}$ is the mutual information between nodes $i$ and $j$. In our case, ${I(v_i;v_j)} = -d_{ij}$, the geometric distances between the observed nodes, which means that the CL-grouping algorithm proposes latent nodes to combine the most correlated pairs into a sibling group, where this group is then represented by a latent node. From the edges between the latent and observed nodes, the CL-grouping algorithm assigns the minimum spanning tree structure for each sub-tree mixture from all possible structures $\mathcal{T}$.

Now, different mixtures for each sub-tree are built. Next, the roots of these mixtures are connected together as children for another latent node. This latent node will hold all messages that are traversing through structure of the mixture to handle the connection between the different sub-trees. The output is a tree model with a set of sub-trees for the limbs' nodes, which are connected together through latent nodes.

This process of constructing different prior structures for each sub-tree is summarised in Figure \ref{MOST::fig_subtree-structure}, where the K-means clustering is employed to categorise the input samples into groups. The appearance and geometric distances of the nodes in a sub-tree with respect to their root are measured and concatenated to build the distance matrix for each group. This matrix is assigned to the CL-Grouping algorithm to determine the structure of a group of samples and this structure defines the parent-child connections between the parts. It forms the configuration prior for a Mixture of the Sub-Tree corresponding to each group of samples.
%
%
\subsection{Mixtures of the Sub-Trees}
\begin{figure*}
\centering
\begin{tabular}{ccccc}
\multirow{-3}{*}{\subfloat[]{\includegraphics[width=1.3in]{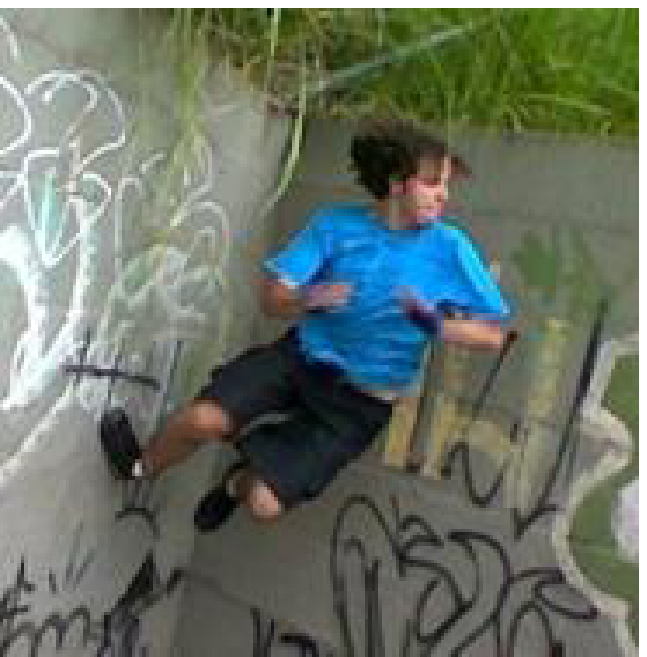}}} &
\multirow{-10}{*}{\subfloat[\label{MOST::fig_hiermodel}]{\includegraphics[width=2.2in]{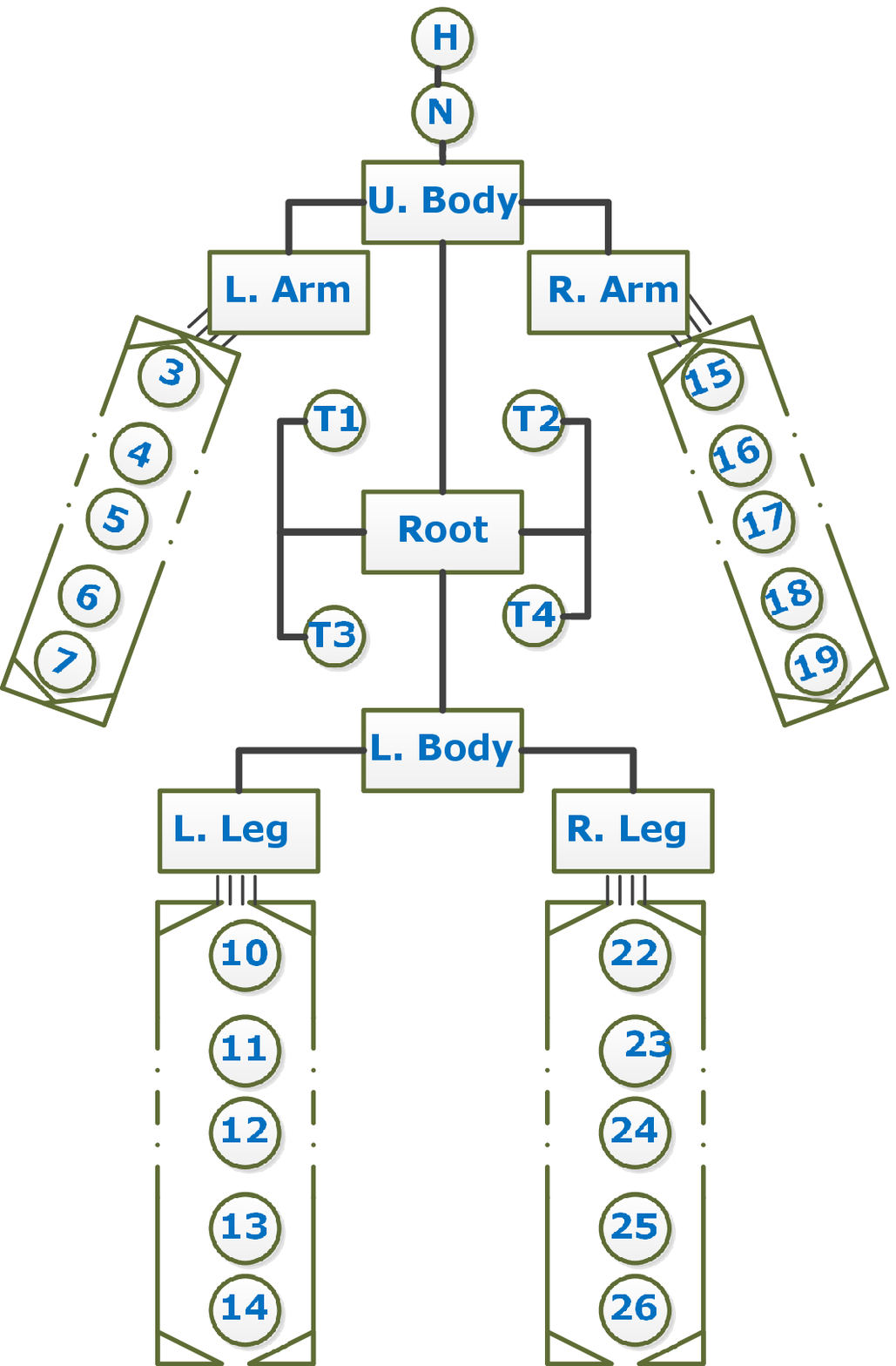}}} &
\subfloat[]{\includegraphics[width = 0.7in,height = 1.5in]{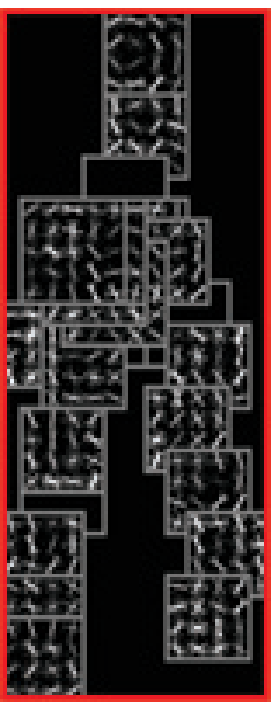}} & 
\subfloat[]{\includegraphics[width = 0.7in,height = 1.5in]{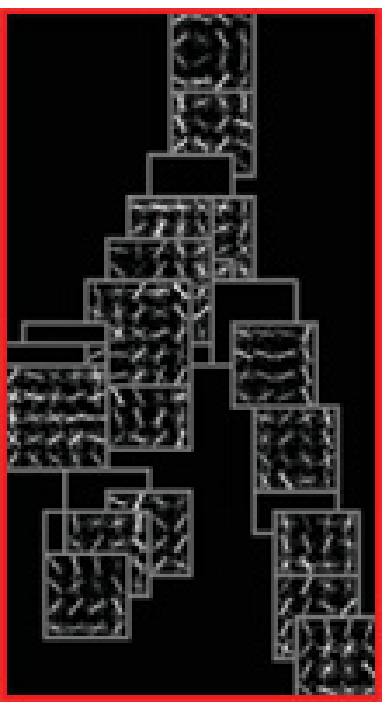}} & 
\subfloat[]{\includegraphics[width=1.2in]{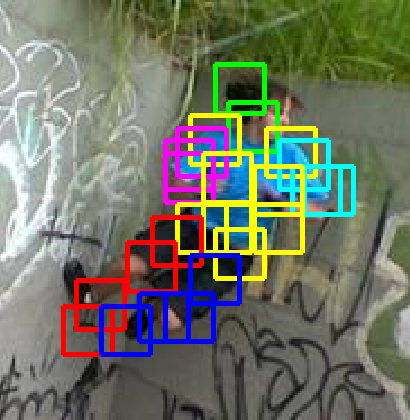}} \\
&&\subfloat[]{\includegraphics[width = 0.7in,height = 1.5in]{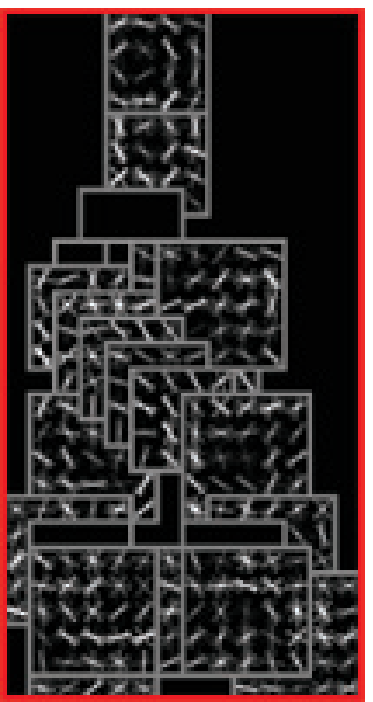}}&
\subfloat[]{\includegraphics[width=0.7in,height = 1.5in]{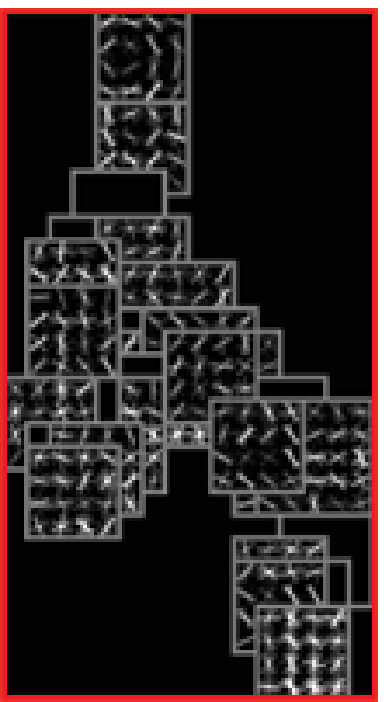}}&
\subfloat[]{\includegraphics[width=1.2in]{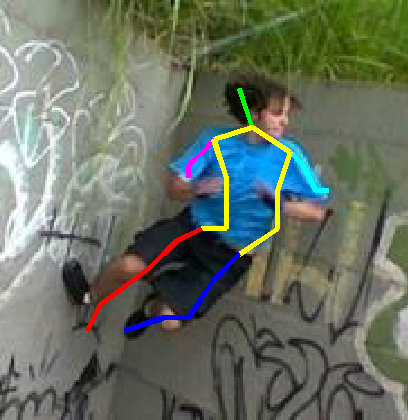}}
\end{tabular}
\caption{\textbf{Mixtures of Sub-Trees model}: (a) Input image. (b) The model in a tree structure where the circles are observed nodes, rectangles represent the latent nodes and the sectional view brackets include the nodes of each sub-tree with no connection on the figure to indicate that the structure of the sub-trees would be determined automatically. (c), (d), (f) and (g) show template-based instances of the model with different mixtures of sub-trees. (e) and (h) show the output pose of applying the model to the input image with the bounding boxes of the parts and the body skeleton overlaid on the image, respectively.}
\label{MOST::fig_model}
\end{figure*}

In this section, the structure of the whole body model is explained. As shown in Figure \ref{MOST::fig_model}, the model has two connected layers to represent the prior of the body pose. The lower layer is for the limbs (legs and arms), which encodes $M$ mixtures for each sub-tree. The upper layer connects the sub-trees' mixtures with the other parts of the torso and the head into a tree model.

Assuming that the configuration of a pose $L$ of an input image $I$ prefers some specific mixtures of the sub-trees $T$, the objective function for estimating that pose configuration is:
\begin{equation}
\label{MOST::eq_pose}
J(T) = \sum_{l \in L^{*}}( \sum_{i \in \textit{mst}(l)} {\phi(T_{i})} + \sum_{i,j \in \textit{most}(l)}{\psi(T_{i},T_{j})}\ )
\end{equation}
where $L^{*}$ are (L.Body, U. Body and Root) latent nodes in Figure \ref{MOST::fig_model}(b),  $\textit{mst}(l)$ is the mixture of sub-trees (children) connected to their latent nodes. The scores of the latent nodes are built from the messages passed to them from the children. Indices $i$ and $j$ represent the mixtures of the trees within each limb.

The first term, $\phi(T_{i})$, holds the potential scores of a whole sub-tree's mixture and the second term is the pairwise scores between two connected sub-trees.  As the number of possible configurations of the model given the mixtures of the sub-trees is large, the unary potentials of a sub-tree are calculated for all mixtures and only the ones with high scores are chosen to be passed to the next layer of the model. This can be formulated via selecting the maximum score values of different mixtures $M$ over $\textbf{v}^*$, which is the set of (observed and latent) nodes of any mixture $T_i$ in the sub-tree. The potential score is calculated as:
\begin{equation}
\label{MOST:eq_subtreescore}
\phi(T_i) = \operatorname*{\arg\,max}_{m \in M} \sum_{i \in \textbf{v}^*}{\phi(I, p_{i}^{m})} + \sum_{i,j}{\phi(I, p_{i}^{m},p_{j}^{m})}
\end{equation}
where $\phi(I,p_{i}^{m})$ is the unary term, which represents the response of convolving the features of location $p$ of image $I$, $\Big(F^{*}(I,p^{m}_k)\Big)$, with a template of type $k$ of part $i$, $\Big(w_{k}(z_{k}^{i})^{T}\Big)$, for mixture $m$ of the underlying sub-tree. The second term in Equation \ref{MOST:eq_subtreescore} represents both the pairwise compatibility and the deformation terms. Type types $z_{k}^{i}$ and $z_{k}^{j}$ of the connected nodes $i$ and $j$ are either being compatible or not, so the compatibility term will be assigned a value 1 or $-\infty$, respectively. The deformation term represents the distance between the expected part location for a specific type and its estimated location. 

At the level of the sub-trees, the unary function will be measured for the nodes of a sub-tree and the mixtures $M$ of a sub-tree. In the upper level of the model, the unary term of the observed nodes in the torso and head (see Figure \ref{MOST::fig_model}) are for the types of these observed nodes.

The second term in Equation \ref{MOST::eq_pose} represents the pairwise scores between the different connected roots of the sub-trees and the other nodes in the models. This again can be calculated via computing both the compatibility and the deformation costs between each pair of sub-trees $T_i$ and $T_j$. In the proposed model, connecting the sub-trees is carried out through latent nodes (see Figure \ref{MOST::fig_model}). The compatibility between the sub-trees has the same meaning as between the nodes in mixture of sub-tree. However, the deformation cost, in the second layer, is the average of all deformation distances between the nodes of the connected sub-trees and can be calculated as:
\begin{equation}
\begin{split}
D(T_i,T_j) = \dfrac{1}{Z} \sum_{v \in {V(T_{i})}} \sum_{k_{i}} -a(z_{k_{i}})d_{i}(v_{k_{i}},\hat{v_{k_{i}}},z_{k_{i}}) + \\
 \sum_{v \in {V(T_j)}} \sum_{k_{j}} -a(z_{k_{j}})d_{j}(v_{k_{j}},\hat{v_{k_{j}}},z_{k_{j}})
\end{split}
\end{equation}
where $Z$ is the number of all mixtures in the two connected sub-trees, $d$ is the squared Euclidean distance between every pair of nodes $(v_{k_{i}})$ and $(\hat{v_{k_{i}}})$ in the sub-tree $T_i$, and $a(z_{k_{i}})$ is a weight parameter for a mixture $k$ of a sub-tree $i$, which needs to be learned from the training data. This weight parameter represents the concatenation the parent-child offsets between the pair of connected nodes in a sub-tree. 

%
%
\subsubsection{Training}
The mixtures of sub-trees model is trained in two steps of learning the parameters. In each of them, the bias, filters and deformation parameters are learnt between the connected nodes in the model. In the first level, the parameters are learned for each mixture of a sub-tree to allow choosing the best fitted limb in the inference phase. The human body pose is first divided into $T$ sub-trees, where each sub-tree has $M$ mixtures with a unique structure. Each mixture of a sub-tree is handled as a tree model with $\vert\textbf{v}^*\vert$ parts, where the instances of each part are clustered into $K$ clusters. The first type of parameters to be learned are $M \times \vert\textbf{v}^*\vert \times K$ filters \{$w$\} for the parts in a sub-tree. The second type of parameters is the compatibility between the nodes, which belong to a specific mixture of sub-tree or between different sub-trees. The deformation weights $a(z_k)$ and the anchor between the connected nodes in a mixture of a sub-tree are the third type of parameters. All of these parameters are initialised by the default values as in \cite{YiYang11}, concatenated in vector $\boldsymbol{w}$, and learned via the max-margin framework and a latent SVM:
\begin{equation}
\label{MOST::EqTraining}
\begin{split}
\operatorname*{\arg\,min}_{\boldsymbol{w}, \xi} \dfrac{1}{2}\sum_{m = 1}^{M} \parallel \boldsymbol{w_m} \parallel_{2} + C \sum_{n} \xi^{n}, \\
s.t.\ y_{i}w_{m_{i}}F^{*}(I,p^{m}_{k_{i}}) \geq (1 - \xi^n),\\
\xi \geq 0, \\
a(z_k) \geq 0
\end{split}
\end{equation}
where $C$ is the regularisation term, which controls the effect of the misclassification on the objective function, $\xi_n$ is the slack variable for sample $n$, and $y_i \in \{+1,-1\}$ represents the labels for either the positive or negative samples.

In the second layer of the model, the same learning scheme as in the first layer is utilised. However, the learned parameters of the first layer are used as initial values for learning the parameters of the whole model in this layer. This initialisation of the parameters accelerates the training process and also positively affects the final result. In this phase of training the model's parameters, the roots of the sub-trees are manipulated as stationary or transitional nodes to measure the overlapping ratios between the connected sub-trees. For this purpose, two other parameters for the upper and lower bounds are learnt, which constrain the overlapping ratios of the potential occluding parts in a pre-trained range.
%
%

\subsubsection{Inference}
As the proposed model has a tree structure, exact inference can be achieved via standard message passing. The pose configuration is obtained via maximising the objective function in Equation \ref{MOST::eq_pose}, which can be performed via sending messages from one node to another one connected in the model. The messages are passed upwards from node $i$ to node $j$ in a mixture $m$ of the sub-tree. Then, the messages are sent from the roots ($r$) of a mixture to the latent root node of its sub-tree ($l$), where the scores of that latent node are initialised to zero and then computed as:
\begin{equation}
\label{MOST::eq_message}
m_{r \rightarrow l}(T_i) = \operatorname*{\max}_{r \in \{roots(M)\}} \phi(T_i, r) \times \theta^{c}(T_i, r)
\end{equation}
where, $\phi(T_i, r)$ is the score of the nodes in a mixture of the sub-tree $T_i$, which is computed with Equation \ref{MOST:eq_subtreescore} for a specific mixture from the $M$ mixtures of the sub-tree. Furthermore, the probability $\theta^{c}(T_i, r)$ determines whether the nodes belonging to the sub-tree are compatible to the mixture type or not. Given an image, the unary terms are calculated via convolving the templates of all the nodes on the mixtures of the sub-trees, the scores are calculated for every node $j$ on the mixtures via message passing as $m_{j}(p_j,z_j) = \sum_{c \in ch{j}} m_{c \rightarrow j}(p_j,z_j)$, see \cite{YiYang11}. Then, the scores of the stationary nodes for each sub-tree are measured using Equation \ref{MOST::eq_message}. 

%
%
\section{Occlusion-Aware Mixture of Sub-Trees (OA-MST)}
\label{MOST::sec_occlusion}

Pictorial structure based methods lack an efficient way to tackle the occlusion problem. Some approaches have been proposed to handle this problem, as mentioned in Section \ref{MOST::sec_literaturereview}. However, these approaches still suffer from the inability to work on both images with and without occlusion and also from the assumption that the occlusion order is known \cite{Sigal2006_short}, \cite{WangM08}, which is generally not the case. 

\cite{WangM08} construct multiple tree models to estimate human pose and to reason the occlusion. In their method, the same strategy as in \cite{Sigal2006_short} is followed via assuming a known order of the body parts and re-weighting the scores of the occluded region through learning binary variables for the occluded and occluding parts. However, detecting the regions with occlusion requires estimating the candidates for the body parts and then re-weighting the scores of the occluded region different from the non-occluded parts. Moreover, reasoning the occlusion between all body parts is computationally expensive when the parts order is unknown and the parts order is impractical. In contrast, an approach is proposed in this paper for an early detection of parts via propagating the messages through the parts in a sub-tree and handling the root of these sub-tree as stationary nodes. Then, the occlusion between these sub-trees is detected and a re-scoring to the scores of the detected regions with occlusion is occurred.

In addition, \cite{AdaptiveChoYL13} developed a method to handle the occlusion order between the body parts relying on the appearance state of the occluding and the occluded parts, where the part, which is more likely to be visible in the overlapping region is handled as occluding and the other one as occluded.

Moreover, in \cite{JiangM08}, a solution to the occlusion reasoning problem has been introduced via penalising the spatially overlapping body parts with an exclusion term to prevent them from appearing in the same body configuration. Their model is a graph with loops, which leads to a non-exact solution, and the root node is also handled as a non-occluded part. Moreover, a regression model is trained from the training data for every pair of parts, which is computationally expensive to be performed for the current PS based models and would require large training datasets, especially when handling each body part with different mixtures. However, the idea of penalising two potentially occluding parts to be in the same spatial position is still valid as the scores coming from the overlapping region between the two parts would be biased and may lead to a wrong configuration pose. 

In this section, an extension to the \emph{MST} approach discussed in Section \ref{MOST::sec_MOST}
is proposed to handle the occlusion occurrence. The main idea is to re-weight the parts with an occlusion region differently than non-occluded regions, without the need to learn a model for each pair of parts as the amount of images with different occlusion modes in the training datasets are not sufficient. This process consists of two steps, occlusion estimation and rectification. 

Detecting the occluding parts in the inference step can be achieved via constraining the search space of the parts needed to be penalised on the overlapping candidates from different parts in different sub-trees. For each part $p$ in a mixture $m$ of a sub-tree $T_{i}$, a potential occlusion set $\mathcal{O}(p_i)$ is constructed, which contains all possible candidates for part $p$ that may be occluded or occluding by other parts in the other sub-trees. The occlusion set is composed initially by applying non-maximum suppression (NMS) on evidence map of the whole mixture $m$ of the sub-tree $T_i$. That is, applying NMS on the level of Mixtures of the Sub-Trees, not on the level of each part. This step will suggest some candidates for each part. Then, the occlusion between the two sub-trees is checked between the candidates in the occlusion set. For a candidate position $u$ of $p_i$, the Intersection over Union (IoU) is utilised to check the overlapping ratio between the bounding boxes around the candidate $u$ and all of the candidates $v$ on the another sub-tree's mixture. An exclusion term will be used:
\begin{equation}
\label{MOST::eq_occlusion}
\psi_{m}(T_{i},p,u) = \log\Big[1 -  \big(\lambda_{ij} * \dfrac{1}{S} \sum_{p \in T_j} \sum_{v \in \mathbb{O}(p_j)} m^{o}_{i \rightarrow j}(p_i,u,p_j,v)\big)\Big]
\end{equation}
where $\lambda_{ij}$ is a weight of how compatible the sub-trees $T_i$ and $T_j$ are (i.e.\ the amount of overlap). This parameter is learnt from the training data and gives more weight to the overlap occurring between the legs and less between legs and arms, for instance. The other term in the equation is the expected value of a candidate $u$ in the sub-tree $T_i$ to be occluding or occluded by any of the candidates of the parts in sub-tree $T_j$, where $S$ is the number of candidate samples of the parts in the sub-tree $T_j$ and $m^{o}_{i \rightarrow j}$ is measured as:
\begin{equation}
\label{MOST:eq_occlusionConstraints}
m^{o}_{i \rightarrow j}(p_i,u,p_j,v) =  \left\{ 
  \begin{array}{l l}
    \theta^{o} = IoU(p_i,p_j) & \  {if \theta^{o} \in \left[L_{ij}, U_{ij}\right]}\\
    0 & \  \text{otherwise.}
  \end{array} \right.
\end{equation}
where $L$ and $U$ are the learnt upper and lower bounds of the possible occlusion ratio between the two sub-trees $T_i$ and $T_j$.

Implementing Equation \ref{MOST::eq_occlusion} for all candidates between the parts of the two sub-trees requires considering all pair computations between the candidates of the parts in the two sub-tree, which results in a huge number of computations, in particular with different mixtures for each sub-tree and for the parts belonging to them. Fortunately, this large number of computations can be drastically reduced, if we only consider the candidates that satisfy two constraints. The first constraint is to consider only candidates that result from applying Non-Maximum Suppression (NMS) on the scores of the root nodes of the underlying sub-trees. Only these candidates and their children in the sub-tree are handled. The second constraint represents the compatibility between the two sub-trees from the upper levels. In other words, only if the two candidates of the nodes of the sub-trees are overlapping with a ratio falling into a pre-trained range (upper and lower bounds).
%
\begin{algorithm}[t]

\DontPrintSemicolon
\SetAlgoLined
\KwIn {$\phi(T_i, m)$, $\phi(T_j) \forall j \in T \setminus \{i\}$, \{$\lambda_{i}$\}, \{$L_i$\}, \{$U_i$\}}
\KwOut{$\hat{\phi}(T_i, m)$ \tcp{\footnotesize{Re-weighted score map}}}
\tcc{\footnotesize{Implement non-maximum suppression to build occlusion list}}
$\mathcal{O}_i \leftarrow NMS(\phi(T_i, m), threshold)$ \;
\For{$p_i \in \mathcal{O}_i$}{
 $\mathcal{O}_j \leftarrow NMS(\phi(T_j, m_j), threshold)\;\;\;\forall m_j \in M(T_j)$\;
  \For{$p_j \in \mathcal{O}_j$}{  
	$\theta^{o} \leftarrow IoU(p_i,p_j)$\;
	Implement Eq. (\ref{MOST:eq_occlusionConstraints}) to get $m^{o}_{i \rightarrow j}$\;
	\If{$\phi(T_i, m,p_j) < \phi(T_j, m_j,p_j)$}{
	$\psi_{m}(T_{i},p,u) \leftarrow $ Apply Equation \ref{MOST::eq_occlusion}\;
   }
  }
  \tcc{\footnotesize{Re-weight only the selected ones}}
$\hat{\phi}(T_i, m,p_i)  \leftarrow \phi(T_i, m,p_i) + max[\psi_{m}(T_{i},p,.)]$\;
}
 \caption{OA-MST algorithm for a mixture $m$ of sub-tree $T_i$}
 \label{MOST::alg_occlusion}
\end{algorithm}

In the proposed occlusion reasoning approach, the exclusion term is added up to the objective function in Equation \ref{MOST::eq_pose}, which re-weights the scores of those points in the intersection area that are between the parts from two different sub-trees. The re-weighting score is applied to the scores of candidates of the occluded parts. Because the occlusion order is unknown, we compare between the score of the two parts and choose the one with the lower value to be the occluded part. The occlusion reasoning step for obtaining the re-weighted score of the occluded regions is summarised in Algorithm ~\ref{MOST::alg_occlusion}.
\section{Experiments}
\label{MOST::sec_experiments}
The performance of the proposed method in estimating the human pose from a single image is evaluated on a number of different experiments.
%
%
\subsection{Datasets}
The proposed method in its two forms (with and without occlusion reasoning) is evaluated on three widely used datasets: The Leeds Sport Poses (LSP), \cite{Johnson10}, Image Parsing (IP), \cite{Ramanan2006} and UIUC People datasets \cite{TranF10}. The LSP dataset consists of 2000 images with different poses for individuals who are performing different sports, where the first half of the images is for training and the second half is for testing. The IP dataset is composed of 305 images for people acting different actions with the entire body being visible. The training set consists of the first 100 images and the remaining ones are for testing. In the UIUC dataset, there are 346 images for training and 247 images for testing where the variation in the poses is quite high. The training set for each of them has been mirrored and rotated with small angles to provide more data in building the model structure and its parameters and also to increase the number of samples with occlusion, which are needed to learn the occlusion parameters. The labels of the body parts for all data have been unified to be observer-centric.
%
%
\subsection{Experiment Settings}
In the model, the parent-child relationships are not defined from the beginning. These relationships are determined by investigating the geometric and appearance similarities between the different body parts in a limb. This results in modelling the different variability in the poses efficiently. To achieve this, each limb is handled as a sub-tree, rooted in a latent node. The training data for each limb are clustered into different mixtures, where each mixture holds a unique structure via applying the CL grouping algorithm recursively, as discussed in Section \ref{MOST::sec_MOST}. 

Technically, two types of mixtures are obtained, namely: Mixtures for each sub-tree, which encode the geometric and appearance relationships between the parts of that sub-tree and mixtures, which represent the orientation of each part with respect to its parent. These mixtures are empirically chosen for each dataset. The number of Mixtures of each Sub-Tree is identified for each dataset by applying cross-validation on the first quarter of each set of images in this dataset.
%
%
\subsection{Evaluation on the LSP Dataset}
\begin{table*}[h]
\centering
\caption{Quantitative evaluation on the LSP dataset: A comparison of the performance of the proposed MST methods with seven state-of-the-art algorithms. The evaluation metric is the PCP, where the numbers are the percentages of the correctly detected parts. The first half of the LSP dataset is utilised for the training of the models and the second half is for testing. This split is the same for all methods mentioned in the table except for \cite{Johnson11}, where the authors used 11000 images for training their model. MST models achieves better accuracies than the other approaches.}
\label{MOST::TableLSP}
\begin{tabular}{|c||c|c|c|c|c|c|c|}
\hline
Method          				& Torso	& Head	& U. Leg	& L. Leg	& U. Arm	& L. Arm	& Total	\\ \hline\hline
\cite{Johnson10} & 78.1	& 62.9	& 65.8		& 58.8		& 47.4		& 32.9		& 55.1	\\ 
\hline
\cite{YiYang11}	& 92.6 	& 87.4 	& 66.4		& 57.7		& 50.0 		& 30.4 		& 58.9	\\ 
\hline
\cite{Johnson11}	& 88.1	& 74.6	& 74.5		& 66.5		& 53.7		& 37.5		& 62.7	\\ 
\hline
\cite{TianZN12}	 & \textbf{95.8}	& \textbf{87.8}	& 69.9		& 60.0 		& 51.9		& 32.8		& 61.3	\\ 
\hline
\cite{EichnerF12} & 86.2	& 80.1	& 74.3		& 69.3		& 56.5		& 37.4		& 64.3	\\ 
\hline
\cite{FWang13} & 91.9	& 86.0	& 74.0		& 69.8		& 48.9		& 32.2		& 62.8	\\ 
\hline
\cite{PishchulinAGS13-conditioned}   & 87.5	& 78.1	& 75.7		& 68.0		& 54.2		& 33.9		& 62.9	\\ 
\hline
MST	 & 85.7	& 81.7	&76.5& 67.8 & 59.6 & 37.8 &  65.1  \\ \hline
OA-MST	 & 87.9	& 82.5	&\textbf{80.1}& \textbf{72.5} & \textbf{61.7} & \textbf{39.9} &  \textbf{67.7}  \\ \hline
\end{tabular}
\end{table*}

In order to learn the pairwise parameters between the body parts, a model for each sub-tree is trained and the learnt bias and deformation parameters are used to initialise their corresponding parameters in the full model. This is different from initialising them with zeros or unity as in \cite{YiYang11} or \cite{TianZN12}.

The PCP \cite{Ferrari2008_short} is used as the evaluation metric for evaluating the proposed method in comparison with other approaches. The PCP computes the percentage of the detected body parts. The total score of a pose estimation method is the product of the human detection rate and the PCP.

Table \ref{MOST::TableLSP} compares the performance of the MST and OA-MST algorithms with seven state-of-the-art methods. Generally, the proposed methods achieves greater accuracy than all the other techniques. More specific, the proposed models improve on the other methods for upper and lower arms and upper legs as well, which indicates that representing these parts as mixtures of sub-trees provides more flexibility and efficiency in localising the articulated limbs. Cross-validation on the first 250 images of the dataset led to choosing the number of mixtures for each sub-tree to be 7. This reflects the high variation in the poses in the LSP datasets. Moreover, the regularisation parameter $C$ in Equation \ref{MOST::EqTraining} has been empirically set to be 0.02 in building the sub-tree models and 0.1 for the full pose model.

\begin{figure*}[!ht]
\centering
\includegraphics[width=6.4in]{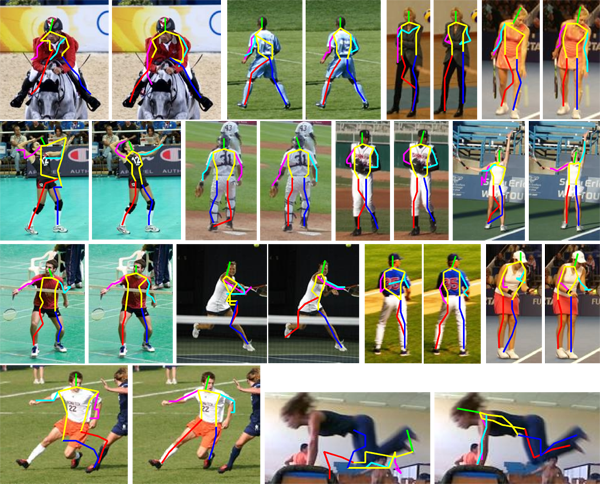}
\caption{Visual comparison on the LSP dataset: Output samples of applying the model of \cite{TianZN12} (left) and the proposed MST method (right), where the highest score poses are overlaid on the input images.}
\label{MOST::fig_LSP-comparison}
\end{figure*}

A visual comparison is shown in Figure \ref{MOST::fig_LSP-comparison} between results of applying the trained model released by the authors of \cite{TianZN12} on the LSP testing samples and the trained MST model without the occlusion handling step. As shown in Figure \ref{MOST::fig_LSP-comparison}, the proposed MST method is superior in accurately obtaining the articulated poses, in particular for the arms and legs.

%
%
\subsection{Evaluation on the IP Dataset}
In the IP dataset, 4 mixtures for each sub-tree are used as the training samples are limited. The MST algorithm is built based on the first 100 images and is tested on the remaining 205 images. Although our proposed method achieves an accuracy of 73.6\%, which is slightly less than the 74.9\% of the state-of-the-art of \cite{YiYang11}, the proposed algorithm is still capable of outperforming it on the detection of the upper and lower arms, as shown in the second last row in Table \ref{MOST::TableIP}. This illustrates the importance of modelling the limbs as mixtures of sub-trees. 
\begin{table*}[h]
\centering
\caption{Quantitative evaluation on the IP dataset: The performance of the proposed MST methods (last two rows) is compared with several state-of-the-art algorithms. The evaluation metric is the PCP.}
\label{MOST::TableIP}
\begin{tabular}{|c|c|c|c|c|c|c|c|}
\hline
Method                     & Torso         & Head          & U. Leg     & L. Leg     & U. Arm     & L. Arm     & Total         \\ \hline\hline
\cite{Andriluka2009}       & 81.4          & 75.6          & 63.2          & 55.1          & 47.6          & 31.7          & 55.2          \\ \hline
\cite{Johnson10}        & 85.4          & 76.1          & 73.4          & 65.4          & 64.7          & 46.9          & 66.2          \\ \hline
\cite{YiYang11} & \textbf{97.6} & \textbf{93.2} & 83.9          & 75.1          & 72.0 & 48.3 & 74.9 \\ \hline
\cite{TianZN12}, 3 Layers & 97.1          & 92.2          & \textbf{85.1} & 76.1 & 71.0          & 45.1          & 74.4          \\ \hline
\cite{TianZN12}, 4 Layers & 96.1          & 92.7          & 81.2          & 71.0          & 69.5          & 39.0          & 71.0          \\ \hline
MST               &        91.2    & 88.8    & 82.0    & 73.2    & 73.2    & 48.6 & 73.6 \\ \hline
OA-MST    & 92.0    & 89.3    & 84.6    & \textbf{77.1}    & \textbf{74.1} & \textbf{50.2}    & \textbf{75.4}\\ 
\hline
\end{tabular}
\end{table*}

After analysing the resulting poses of applying the MST algorithm on the IP dataset, the need for rectifying the errors in the detections of the body limbs, which happened due to self-occlusion, became very clear. One more model is learnt to tackle the problem of double counting and self-occlusion in the samples of this dataset. The learnt model follows the steps in OA-MST Algorithm ~\ref{MOST::alg_occlusion} to identify the occluded regions and to re-weight their scores based on the overlapping ratios. 

Because the scores of the regions are only representing the likelihood of a body part to be in this region, a large number of potential overlapping regions between the different body parts is needed. This leads to a high computational cost to determine the part regions to be re-weighted. The number of overlapping regions is drastically decreased by applying NMS on the roots of each sub-tree and considering only the scores of parts in two different sub-trees, which are sharing the same parent locations. This trick handles only the regions that have a high impact on the occlusion occurrences between the body parts.

The results in the last row of Table \ref{MOST::TableIP} show the performance of rectifying the poses with the occlusion-aware step proposed in Section \ref{MOST::sec_occlusion}. The total PCP score of the detected parts is (75.4\%), which outperforms all other approaches for this dataset.

\begin{figure*}[!ht]
\centering
\includegraphics[width=6.4in]{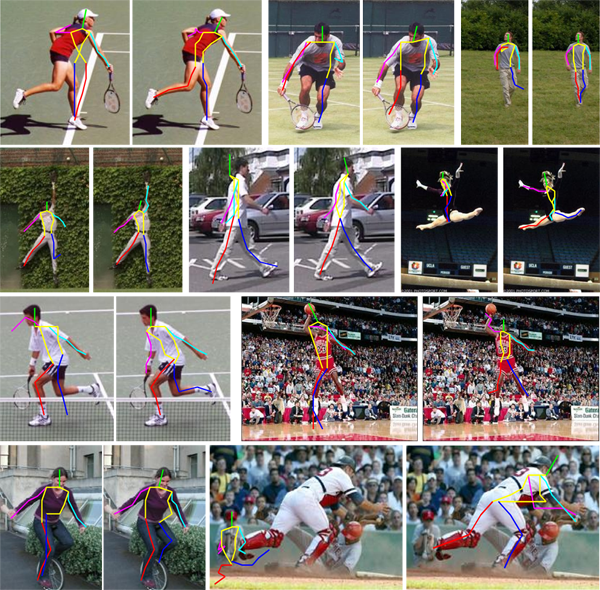}
\caption{Visual comparison on the IP dataset: Sample results of applying the constructed model in \cite{YiYang11} (left) and the proposed MST model (right). In this experiment, the occlusion awareness to the MST model is added, which leads to better results, especially for the body limbs on the majority of the input images.}
\label{MOST::fig_IP-comparison}
\end{figure*}

A qualitative comparison is provided in Figure \ref{MOST::fig_IP-comparison} between the output of the method in \cite{YiYang11} and the proposed MST approach. For the detection of limbs, it is clear that the performance of the MST method is superior.
%

%
%
\subsection{Cross Database Evaluation}
\begin{table*}[!ht]
\centering
\caption{Quantitative evaluation on the UIUC dataset: PCP comparison between the proposed MST approaches and three other techniques. In the MST approaches, we learn the model on the training set of the LSP dataset and evaluate on the testing samples of the UIUC dataset.}
\label{MOST::TableUIUC}
\begin{tabular}{|c|c|c|c|c|c|c|c|}
\hline
Method                     & Torso         & Head          & U. Leg     & L. Leg     & U. Arm     & L. Arm     & Total         \\ \hline\hline
\cite{Andriluka2009}        &  88.3          & 81.8          & 64.0          & 50.6          & 42.3          & 21.3          & 52.6          \\ \hline
\cite{Wang2011Poselets}         & 86.6          & 68.8          & 56.3          &  50.2          & 30.8          &  20.3          & 47.0          \\ \hline
\cite{PishchulinAGS13-conditioned} & \textbf{91.5} & 85.0 & 66.8          & 54.7          & 38.3 & 23.9 & 54.4 \\ \hline
MST      &  83.8   &    83.8   & 73.9 & 60.1 & 49.0 & 35.2 &60.4 \\ \hline    
OA-MST      &  86.2   &    \textbf{86.4}   & \textbf{77.9} & \textbf{65.5} & \textbf{52.3} & \textbf{38.1} &\textbf{63.6} \\ \hline    
\end{tabular}
\end{table*}

For evaluating the performance of the proposed algorithm across datasets, this experiment is performed on the UIUC dataset, where the model is built on the training data of the LSP dataset. In Table \ref{MOST::TableUIUC}, the results of this experiment are shown together with some other known results on the dataset from the literature. The MST algorithms provide superior results compared to the other approaches, in particular with the limbs. This indicates the quality of the proposed algorithms in modelling and detecting the parts in the limbs. Regarding the generalisation of the proposed algorithms, the MST methods produce better results across datasets.
\begin{figure}[h]
\centering
\begin{tabular}{c}
\subfloat[]{\includegraphics[width = 0.45\textwidth]{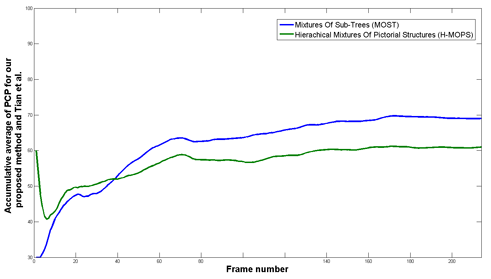}} \\
\subfloat[]{\includegraphics[width = 0.45\textwidth]{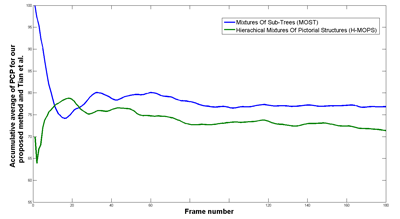}}
\end{tabular}
\caption{\textbf{Evaluation on the KTH Multi-view Football dataset:} The MST model and the model from \cite{TianZN12} are applied, PCP values are calculated and the averages to the current frame are visualised for sequences (a \& b) from the dataset in \cite{kazemi2013multi}. The evaluation is on a frame-by-frame basis.}
\label{MOST::fig_2dsport}
\end{figure}

\cite{Wang2011Poselets} have built a model on the UIUC training data and evaluated on the UIUC testing data. In \cite{Andriluka2009} and \cite{PishchulinAGS13-conditioned}, the parameters of the model have been learnt on both the training data of UIUC and LSP. In contrast, in the experiment here, the parameters of the proposed model are constructed based solely on the LSP dataset. With these settings, the OA-MST algorithm is still capable of outperforming the other approaches and achieving 9\% (absolute) accuracy higher than the closest approach.

Additionally, the MST algorithm is evaluated on two different video sequences for two players from the KTH Multiview Football dataset \cite{kazemi2013multi}. The same model is applied to the video clips and evaluated using the average of the accumulative PCP, as shown in Figure \ref{MOST::fig_2dsport}. The MST approach outperforms the results in \cite{TianZN12} in the two sequences, where the PCP is calculated on a frame-by-frame basis. A detailed description of this experiment and the comparison on all frames can be found in the supplementary material.

From the experiments, which are conducted on the three datasets, it is shown that the proposed MST algorithm has the ability to localise the body parts accurately and to provide an accurate pose estimation. Moreover, providing a solution to the problem of occlusion without losing the benefit of representing the model as a tree improves the results and rectifies the location of the parts efficiently.
%
%
\section{Conclusions}
\label{MOST::sec_conclusion}
This paper addresses the problem of modelling the apparent non-physical connection due to self-occlusion via representing the body limbs as mixtures of sub-trees and learning the pairwise relations in a tree model. In addition, an occlusion handling process to reduce the erroneous detections is proposed, which may occur because of self-occlusion. From the experiments, we conclude that our method improves the pose prediction accuracy over the results reported on three different datasets. 

Furthermore, we notice that structuring the body parts based on the appearance and geometric differences results in a better chance for the part to be located in the right mixture. Moreover, determining the parent-child relationships based on the spatial and appearance correlation between the body parts resulted in learning the pairwise configurations more efficiently, contrary to what is the case in the standard mixtures of body part based methods. 

In the future, a further analysis of the reduced accuracy in the head and torso parts is needed to be conducted and rectified. Moreover, we will extend the OA-MST algorithm to predict the poses of multiple persons with inter-occlusion and different interactions. This can help recognising human activities in videos.
\ifCLASSOPTIONcaptionsoff
  \newpage
\fi
\bibliographystyle{IEEEtran}
\bibliography{Ibrahimlong}
\end{document}